\documentclass{article}

\usepackage[final,nonatbib]{neurips_2022}

\usepackage[utf8]{inputenc} 
\usepackage[T1]{fontenc}    
\usepackage{hyperref}       
\usepackage{url}            
\usepackage{booktabs}       
\usepackage{amsfonts}       
\usepackage{nicefrac}       
\usepackage{microtype}      
\usepackage{xcolor}         

\usepackage{graphicx}

\usepackage{array}
\newcolumntype{P}[1]{>{\centering\arraybackslash}p{#1}}

\usepackage{floatrow}
\newfloatcommand{capbtabbox}{table}[][\FBwidth]

\usepackage{amsmath}
\usepackage{amssymb}
\usepackage{comment}

\title{Few-Shot Continual Active Learning by a Robot}

\author{%
  Ali~Ayub
    \\
  University of Waterloo\\
  Waterloo, ON N2L3G1, Canada \\
  \texttt{a9ayub@uwaterloo.ca} \\
  \And
  Carter Fendley \\
  Capital One \\
  New York, NY 10017, USA \\
  \texttt{ ccf5164@psu.edu} \\
}

\begin{document}

\maketitle

\begin{abstract}
In this paper, we consider a challenging but realistic continual learning problem, Few-Shot Continual Active Learning (FoCAL), where a CL agent is provided with unlabeled data for a new or a previously learned task in each increment and the agent only has limited labeling budget available. Towards this, we build on the continual learning and active learning literature and develop a framework that can allow a CL agent to continually learn new object classes from a few labeled training examples. Our framework represents each object class using a uniform Gaussian mixture model (GMM) and uses pseudo-rehearsal to mitigate catastrophic forgetting. The framework also uses uncertainty measures on the Gaussian representations of the previously learned classes to find the most informative samples to be labeled in an increment. We evaluate our approach on the CORe-50 dataset and on a real humanoid robot for the object classification task. The results show that our approach not only produces state-of-the-art results on the dataset but also allows a real robot to continually learn unseen objects in a real environment with limited labeling supervision provided by its user\footnote{Preliminary ideas \cite{ayub_icra_workshop_2021,ayub_roman_workshop_2021} related to this work were presented at workshops in RoMan 2020 and ICRA 2021.}.
\end{abstract}

\section{Introduction}
\label{sec:introduction}
\noindent Continual learning (CL) \cite{Rebuffi_2017_CVPR,kirkpatrick17,Ayub_IROS_20, ayub2021eec} has emerged as a popular area of research in recent years because of its limitless real-world applications, such as domestic robots, autonomous cars, etc. Most continual machine learning models \cite{Hou_2019_CVPR,Wu_2019_CVPR,Tao_2020_CVPR,Ayub_2020_CVPR_Workshops}, however, are developed for constrained task-based continual learning setups, where a CL model continually learns a sequence of tasks, one at a time, with all the data of the current task labeled and available in an increment. Real world systems, particularly autonomous robots, do not have the luxury of getting a large amount of labeled data for each task. In contrast, robots operating in real-world environments mostly have to learn from supervision provided by their users \cite{Dehghan19,Ayub_IROS_20,ayub_icsr_2020}. Human teachers, however, would be unwilling answer a large number of questions or label a large amount of data for the robot. It would therefore be useful for robots to self-supervise their learning, and ask the human teachers to label the most informative training samples from the environment, in each increment. In this paper, we focus on this challenging problem, termed as Few-Shot Continual Active Learning (FoCAL).

One of the main problems faced by continual machine learning models is catastrophic forgetting, in which the CL model forgets the previous learned tasks when learning new knowledge. In recent years, several works in CL have focused on mitigating the catastrophic forgetting problem \cite{french19,Rebuffi_2017_CVPR,ayub2021eec}. Most of these works, however, are developed for the task-based continual learning setup, where the model assumes that all the data for a task is available in an increment and it is fully labeled. These constraints are costly and limit the real-world application of CL models on robots. Active learning has emerged as an area of research in recent years, where machine learning models can choose the most informative samples to be labeled from a large corpus of unlabeled data, thus reducing the labelling effort \cite{Siddiqui_2020_CVPR,Yoo_2019_CVPR}. Most active learning techniques use uncertainty sampling to request labels for the most uncertain objects \cite{Yoo_2019_CVPR,Siddiqui_2020_CVPR,singh20}. These techniques, however, do not learn continually and thus would suffer from catastrophic forgetting. These issues related to the development of "close-world`` techniques for continual learning, active learning and open-set recognition have been explored in detail in \cite{mudt2020}.  

In this paper, we consider FoCAL for the online continual learning scenario for the image classification task. In this setup, a CL model (applied on a robot) receives a small amount of unlabeled image data of objects from the environment in an increment, where the objects can belong to the previously learned classes by the model, or new classes. The model is allowed to get a small number of object samples to be labeled by the user. As the model continues to learn from new training samples, it does not have access to the raw image data of the previously learned objects. Overall, FoCAL is a combination of multiple challenging problems in machine learning, mainly Few-Shot Class Incremental Learning (FSCIL) \cite{Tao_2020_CVPR,Ayub_2020_CVPR_Workshops}, Active Learning \cite{Siddiqui_2020_CVPR,Gal17}, and online continual learning \cite{Aljundi_2019_CVPR}. To solve FoCAL, we get inspiration from the continual learning and active learning literature, to develop protocols for continual learning models so that they can actively choose informative samples in an increment. Particularly, we take inspiration from FSCIL literature to develop a new FoCAL model, in which we learn and preserve the feature representation of the previously learned objects classes by modelling them as Gaussian mixture models. To mitigate catastrophic forgetting, we use pseudo-rehearsal \cite{Robins95} using the samples generated from the Gaussian distributions of the old classes, thus removing the need to store raw data for the classes. Further, to choose most informative samples from an unlabeled set, we use a combination of predictive entropy \cite{shannon_1949,Gal17} and viewpoint consistency metrics \cite{Siddiqui_2020_CVPR,singh20} on the GMM representation of the previously learned classes. We perform extensive evaluations of our proposed approach on the CORe-50 dataset \cite{lomonaco17}, and on a real humanoid robot in an indoor environment. Our approach outperforms state-of-the-art (SOTA) continual learning approaches for FoCAL on the CORe-50 datast with significant margins. Further, our approach can also be integrated on a humanoid robot, and allow the robot to learn a large number of common household objects over a long period of time with limited supervision provided by the user. Finally, as a part of this work, we also release the object dataset collected by our robot as a benchmark for future evaluations for FoCAL (available here: \url{https://tinyurl.com/2vuwv8ye}). 

\begin{figure*}
\centering
\includegraphics[width=1.0\linewidth]{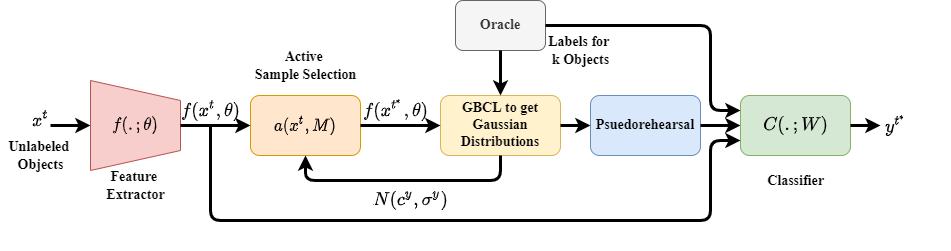}
\caption{\small Our overall framework for FoCAL. In each increment $t$, the features extracted for unlabeled objects $f(x^t;\theta)$ are passed through the acquisition function $a(x^t,\mathcal{M})$ to get $k$ most informative samples $x^{t^*}$, which are labeled by the oracle. The labeled feature vectors are used to update the GMM representation of the learned classes $Y^t$. Pseudo-rehearsal is used to replay old class data, and the classifier model $C(.;W)$ is trained on the pseudo-samples of the old classes and the labeled feature vectors in the $t$th increment.
}
\label{fig:framework}
\end{figure*}

\section{Few-Shot Continual Active Learning}
\label{sec:methodology}
\noindent We define the Few-Shot Continual Active Learning (FoCAL) problem as follows: Suppose that an AI agent (e.g. a robot) gets a stream of unlabeled data sets $D_{pool}^1, D_{pool}^2, ...,D_{pool}^t,...$ over $t$ increments, where $D_{pool}^t = \{x^t_i\}_{i=1}^{|D_{pool}^t|}$. In each increment, a continual learning model $\mathcal{M}$ with parameters $\Theta$ can only obtain a small number ($k^t<|D_{pool}^t|$) of samples to be labeled. Given the model $\mathcal{M}$, an acquisition function $a(x^t,\mathcal{M})$, where $x^t \in D_{pool}^t$, is used by the AI agent to find the most informative samples to be labeled in an increment $t$: $x^{t^*} = {\rm{argmax}}_{D^t_{pool}} a(x^t,M)$.
Therefore, in each increment $t$, the model $\mathcal{M}$ gets trained on small subsets of labeled data $D^t = \{(x^{t^*}_i,y_i^t)\}_{i=1}^{|k^t|}$, where $y_i^t \in Y^t$ represents the class label of $x_i^{t^*}$ and $Y^t$ is the set of classes in the $t$-th increment. Note that unlike most continual learning setups, $Y^i \cap Y^j \neq \varnothing$, $\forall i\neq j$. After training on $D^t$, the model $\mathcal{M}$ is tested to recognize all the encountered classes so far $Y^1, Y^2, ..., Y^t$. The main challenges of FoCAL are three-fold: (1) avoid catastrophic forgetting, (2) prevent overfitting on the few training samples, (3) efficiently choose most informative samples in each increment. 

For FoCAL for the task of object classification, we consider the model $\mathcal{M}$ (a CNN) as a composition of a feature extractor $f(.;\theta)$ with parameters $\theta$ and a classification model with weights $W$. The feature extractor transforms the input images into a feature space $\mathcal{F}\in\mathbb{R}^n$. The classification model takes the features generated by the feature extractor, and generates an output vector followed by a softmax function to generate multi-class probabilities. In this paper, we use a pre-trained feature extractor, therefore parameters $\theta$ are fixed. Thus, we incrementally finetune the classification model on $D^1, D^2, ...$ and get parameters $W^1, W^2, ...$. In an increment $t$, we expand the output layer by $|Y^t|$ neurons to incorporate new classes. Note that this setup does not alleviate the three challenges of FoCAL mentioned above. The subsections below describe the main components of our framework (Figure \ref{fig:framework}) to transform this setup for FoCAL. 

\subsection{GMM Based Continual Learning (GBCL)}
\label{sec:gbcl}
We aim to develop a model that not only helps the system with continual learning but is also motivated by the newness of an object. To accomplish this, we must evaluate how different an incoming object is from previously learned object classes, ideally without any additional supervision. To accomplish this, we consider a clustering-based approach to represent the distribution of object classes. Unlike previous works on clustering-based approaches for continual learning \cite{Ayub_2020_CVPR_Workshops,Dehghan19} that represent the object classes as mean feature vectors (centroids), we estimate the distribution of the each object class using a uniform Gaussian mixture model (GMM). We believe that representing each class data as a GMM may better represent the true distribution of the data rather than assuming that the distribution is circular. We call our complete algorithm for continually learning GMMs of multiple object classes as GMM based continual learning (GBCL).

Once the $k$ feature vectors ($D^t$) selected by acquisition function (Section \ref{sec:curiosity_module}) as most informative samples are labeled by the oracle in increment $t$, GBCL is applied to learn GMMs for the classes $Y^t$. For each $i$th feature vector $x_i^{t^*}$ in $D^t$ labeled as $y_i^t$, if $y_i^t$ is a new class never seen by the model before, we initialize a new Gaussian distribution $\mathcal{N}(x_i^t,O)$ for class $y$ with $x_i^t$ as the mean (centroid) and a zero matrix ($O$) as the covariance matrix\footnote{We do not describe mixing coefficients here, as they will always be $1/n$ for a uniform GMM, where $n$ is the number of mixture components.}. Otherwise, if $y_i^t$ is a known class, we find the probabilities $\mathcal{N}(x_i^t|c_1^y,\sigma_1^y), ...,\mathcal{N}(x_i^t|c_j^y,\sigma_j^y), ..., \mathcal{N}(x_i^t|c_{n^y}^y,\sigma_{n^y}^y)$ for $x_i^y$ to belong to all the previously learned Gaussian distributions for class $y$, where $n^y$ is the total number of mixture components in the GMM for class $y$, and $c_j^y$ and $\sigma_j^y$ represent the centroid and covariance matrix for the $j$th mixture component of class $y$, respectively. If the maximum probability among the calculated probabilities for all the distributions is higher than a pre-defined probability threshold $P$, $x_i^t$ is used to update the parameters (centroid and covariance matrix) of the most probable distribution ($\mathcal{N}(c_j^y,\sigma_j^y)$) in class $y$. The updated centroid $\hat{c}_j^y$ is calculated as a weighted mean between the previous centroid $c_j^y$ and $x_i^t$:

\begin{equation}
    {\hat{c}}_j^y = \frac{w_j^y \times c_j^y + x_i^t}{w_j^y+1}
\end{equation}

\noindent where, $w_j^y$ is the number of images already clustered in the $j$th (most probable) Gaussian distribution. The updated covariance matrix $\hat{\sigma}_j^y$ is calculated based on the procedure described in \cite{dasgupta07}):

\begin{equation}
    \hat{\sigma}_j^y = \frac{w_j^y -1 }{w_j^y}\sigma_j^y + \frac{w_j^y - 1}{{w_j^y}^2}(x_i^t-\hat{c_j^y})^T(x_i^t-\hat{c}_j^y)
\end{equation}

\noindent where, $\sigma_j^y$ is the previous covariance matrix and $(x_i^t - \hat(c)_j^y)^T(x_i^t - \hat{c}_j^y)$ represents the covariance between $x_i^t$ and $\hat{c}_j^y$. If, on the other hand, the maximum probability among the calculated probabilities for all the distributions is lower than $P$, a new Gaussian distribution $\mathcal{N}(x_i^t,O)$ is created for class $y$ with $x_i^t$ as the centroid and $O$ as the covariance matrix. 

The result of this process is a set of $N^t$ uniform GMMs with parameters $\phi^1, \phi^2, ..., \phi^{N_t}$ for $N^t$ classes learned up till increment $t$. Note that instead of using the number of mixture components as a hyperparameter, we use the probability threshold. This way we can start with a simple distribution model for each class (a single mixture component) and add more mixture components only when the new images of the class are too dissimilar from the previous mixture components, and thus cannot be modeled by the GMM. Therefore, the total number of mixture components for each class can be different based on the similarity between the images of the class. In section \ref{sec:curiosity_module}, we use the same idea of dissimilarity between an unlabeled image and a GMM to predict most informative samples.

\subsubsection{Pseudo-rehearsal and Classifier Training}
To avoid catastrophic forgetting, we use pseudo-rehearsal \cite{Robins95} to replay the old classes when learning from new data in increment $t$. For pseudo-rehearsal, we sample the Gaussian distributions in the GMMs of all the previously learned classes to generate a set of pseudo-feature vectors. Note that we also store the total number of images clustered in each Gaussian distribution of the classes ($w_j^y$) during the GMM learning phase (Section \ref{sec:gbcl}). Therefore, we generate the same number of pseudo-feature vectors as the original number of images for each class. After generating the pseudo-feature vectors, the classifier model $C(.;W)$ is trained using the labeled dataset $D^t$ in increment $t$, and the pseudo-feature vectors of the previous classes.

For classification of a test image $x$, it is first passed through the feature extractor $f(x,\theta)$ and then through the classifier $C(f(x,\theta),W)$. Softmax function ($\sigma$) is then applied on the output to generate class probabilities, and the class $y^*$ with the maximum probability is predicted as the label for the test image: $y^* = {\rm{argmax}}_y \; \sigma(W^Tf(x,\theta))$.


\subsection{Active Learning using GMMs}
\label{sec:curiosity_module}
We quantify the novelty of an object in terms of how much the model is uncertain about the object. Unlike most active learning setups \cite{Gal17,Siddiqui_2020_CVPR}, in FoCAL the model does not have access to a training set in each increment for training the model to predict uncertain object classes. Further, even if the model does get trained to predict unknown object classes in the first increment, it will catastrophically forget the criterion of novelty as it continually learns new object classes in the subsequent increments (unknown classes in the first increment become known to the model in the subsequent increments). Therefore, we do not train our model 
for active learning, and instead use the GMM representations of the previously learned object classes to predict the most uncertain objects.

Considering the FoCAL setup (as described in Section \ref{sec:methodology}), in an increment $t$, the AI agent gets an unlabeled dataset $D_{pool}^t$ and it must find $k<|D_{pool}^t|$ most informative object samples from the dataset to be labeled. To develop an acquisition function for this, we use a combination of two active learning techniques applied to the GMM representations of the previously learned object classes.

First, we use the prediction entropy $\mathbb{H}[y^*|x_i^t]$ of an object as the acquisition function \cite{shannon_1949}:

\begin{equation}
    \mathbb{H}[y^*|x_i^t] = - \sum_{y=1}^{N^{t-1}} p(y^*=y|x_i^t) {\rm{log}}p(y^*=y|x_i^t)
\end{equation}

For an unlabeled data point $x_i^t \in D_{pool}^t$, we find the predictive probability of $x_i^t$ using the GMM representation of the object classes learned in the previous increments. The predictive probability $p(x_i^t|\phi^y)$ for of $x_i^t$ to belong to the GMM of a class $y$ can be defined as:

\begin{equation}
    p(x_i^t|\phi^y) = \frac{1}{n^y} \sum_{j=1}^{n^y} \mathcal{N}(x_i^t|c_j^y,\sigma_j^y) 
\end{equation}

Intuitively, if a sample $x_i^t$ is already learned by the AI agent, then its probability to belong to one of the previously learned class GMMs must be high, and thus entropy for $x_i^t$ must be low. Therefore, top $k$ samples with the highest entropy can be chosen as the most informative samples.

For the second technique, we use the idea of viewpoint consistency used in active learning \cite{Siddiqui_2020_CVPR,singh20}. The main idea is that if an object is already learned by the agent before, then consistent predictions must be produced for the object under different viewpoints (see Figure \ref{fig:multiple_views}). 
We again use the GMM representations of the previously learned classes for this acquisition function. Let's consider that there are multiple ($l$) viewpoints $x_1^t, ..., x_i^t, ..., x_{l}^t$ of an unlabeled object $j$ available to the AI agent in increment $t$. We find the predictive probability of each $x_i^t$ to belong to the GMM representations of the previously learned classes (equation (6)). Using the maximum predictive probability over all the classes we find the class prediction $y_i^{t^*}$ for each $x_i^t$ as $y_i^{t^*}={\rm{argmax}}_{y=1,...,N^t} p(x_i^t|\phi^y)$. Next, we count the total number of times each of the $N^{t-1}$ classes is predicted among the different viewpoints of the object $j$. Let $S_j = \{s_j^1, ..., s_j^y,...,s_j^{N^{t-1}}\}$ represent the total number of times each class is predicted among the different viewpoints of the object. We normalize set $S_j$ by dividing each $s_j^y$ by $l$ such that $\sum_{y=1}^{N^{t-1}} s_j^y = 1$. We then take the inverse of the maximum value in $S_j$ ($1/{\rm{max}}S_j$), which represents the inverse of the highest percentage of viewpoints of object $j$ that are predicted consistently. Thus, we use this term as the inconsistency score of the object $j$.

We use a combination of predictive entropy and viewpoint consistency to generate the final acquisition function for our framework. Before combining the two metrics, we transform the predictive entropy for multiple viewpoints of the same objects as it is originally designed for individual data samples. For multiple viewpoints of an object $j$, we find the predictive entropy of all the viewpoints individually and then take an average to get the overall predictive entropy of object $j$. Therefore, we maximize the following combined function to get the most informative samples in increment $t$:

\begin{equation}
    \frac{\delta}{l} \sum_{m=1}^{l} \mathbb{H}[y^*|x_m^t] + (1-\delta) \frac{1}{{\rm{max}}S_j} 
\end{equation}

\noindent where, $\delta$ is a hyperparameter that controls the contribution of predictive entropy and viewpoint consistency towards the overall uncertainty score of an object. Note that the viewpoint consistency can only be applied when the agent has access to multiple viewpoints of the same object. In cases when multiple viewpoints are unavailable, we can use the predictive entropy alone. In our experiments (Section \ref{sec:experiments}), however, we use multiple viewpoints of all the objects.


\section{Experiments}
\label{sec:experiments}
\noindent We first evaluate our approach for FoCAL on the Core-50 dataset \cite{lomonaco17} and then using the Pepper robot. We begin by presenting the implementation details and then compare our approach against SOTA continual learning approaches on the CORe-50 dataset. Finally, we present the evaluations of our approach on Pepper. Details about the CORe-50 dataset are in Section \ref{sec:core50}.

\subsection{Implementation Details}
We used the Pytorch deep learning framework \cite{torch19} for implementation and training of all neural network models. We used ResNet-18 \cite{He_2016_CVPR} pre-trained on the ImageNet dataset \cite{Krizhevsky12} as a feature extractor for GBCL. The same pre-trained network was also used in all the other continual learning approaches for a fair comparison. For GBCL, we only used the diagonal entries of the covariance matrix to keep the memory budget from growing drastically.

For FoCAL experiments on the CORe-50 dataset, we develop a new experimental setup, as the standard active learning and continual learning experimental setups are not sufficient to test FoCAL (see Section \ref{sec:comparison_with_setups} for more details). In this experimental setup for FoCAL on CORe-50, we combined all 8 training sessions in CORe-50 to generate 400 training object instances. In each increment, we randomly sampled $m=5$ object instances (from 400 instances) and allowed the model to learn the label of $k=1$ out of 5 object instances, making it a challenging problem to learn from a single object instance in each increment. Once an object was learned by the model, it was removed from the complete set of objects and, thus, was not available to the model in later increments. Hence, we allowed the model to learn all objects in 400 increments with one object learned in each increment. We used the test set accuracy at each increment as the evaluation metric. After training in each increment, the model was tested on the complete test set of the CORe-50 dataset. We only report the accuracy for 70 increments for this experiment because all the approaches learn all 10 classes by 70 increments and the final accuracy starts to saturate.

For robustness, all the experiments were performed 5 times with random seeds. We report the average and standard deviation of the accuracies. Hyperparameters P and $\delta$ were chosen using cross-validation and were set to 0.2 and 0.7, respectively for all increments. For the shallow neural network used for classification in GBCL, we used a linear layer of the same size as the feature vectors generated by the ResNet (512$\times$1). The shallow network was trained for 25 epochs using the cross-entropy loss optimized with stochastic gradient descent (with 0.9 as momentum). A fixed learning rate of 0.01 and minibatches of size 64 were used for training. 

\subsection{Experiments on the CORe-50 Dataset}
\label{sec:core50_experiment}
We compare our approach against 6 continual learning approaches (LWF \cite{Li18}, EWC \cite{kirkpatrick17}, CWR \cite{lomonaco17}, iCaRL \cite{Rebuffi_2017_CVPR}, NCM \cite{Dehghan19}, CBCL \cite{Ayub_2020_CVPR_Workshops}), finetuning (FT) and a few-shot batch learning baseline (FLB) \cite{Chen19}. FLB uses a pre-trained CNN to extract feature vectors for images and trains a linear classifier using cross entropy loss. FLB is trained on the complete training data of the previous increments and the current increment. In other words, FLB does not learn continually and therefore should have an advantage over other continual learning approaches (including ours). FT uses the same architecture as FLB but FT is trained only on the data of the current increment. FT suffers from catastrophic forgetting and therefore should produce lower accuracy than other continual learning approaches. To the best of our knowledge, none of the eight approaches can be directly applied to FoCAL. Hence, for FLB, FT, LWF, EWC and CWR we used softmax-based uncertainty sampling to find most uncertain objects. For the softmax-based uncertainty scores, we found the softmax output of each image for an object instance. Then, we took the average of the maximum probability in the softmax output of each image of the object. The object with the highest average probability score was chosen to be labeled. Because iCaRL, NCM, and CBCL use centroids for classification, we used the average distance of the images of the new object from the centroids as the uncertainty score. A brief description of all six continual learning approaches is provided in Section \ref{sec:sota_cl_approaches}.

\begin{figure}
\begin{floatrow}
\ffigbox{%
  \includegraphics[scale=0.5]{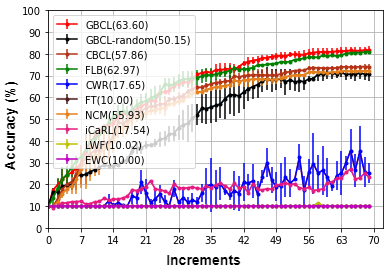}
}{%
  \caption{\small Comparison of our method (red curve) to SOTA approaches in terms of classification accuracy on the Core-50 dataset. Average incremental accuracy is reported in parenthesis. The curves show average and standard deviation of 5 runs with random seeds.}
\label{fig:curiosity_experiment_dataset}
}
\capbtabbox{%
  \begin{tabular}{|P{2.0cm}|P{2.0cm}|}
     \hline
    \textbf{Methods} & \textbf{Increments} \\
    \hline
     FT & 23\\
    \hline
    LWF & 23\\
    \hline
    EWC & 24\\
    \hline
    iCaRL & 18\\
    \hline
    NCM & 17\\
    \hline
    CWR & 21 \\
    \hline
    FLB & 19\\
    \hline
    CBCL & 17\\
    \hline
    GBCL-rand & 22\\
    \hline
    \textbf{GBCL} & \textbf{15}\\
 \hline
 \end{tabular}
}{%
  \caption{\small Total number of increments needed by each method to learn 10 classes in CORe-50.}
  \label{tab:classes}
}
\end{floatrow}
\end{figure}

\subsubsection{Comparison with SOTA Approaches}
Figure \ref{fig:curiosity_experiment_dataset} compares our GBCL approach against random sampling and other SOTA approaches in terms of classification accuracy on the fixed test set for 10 classes. Our approach (GBCL) and FLB produce similar results for all 70 increments and GBCL produces slightly better accuracy than FLB in some increments. FLB, however, uses the data in all the previous increments at each new increment, while our approach only uses the data in the current increment and pseudo-feature vectors of the previous classes. We believe the reason for similar results for the two approaches is that GBCL generates and uses pseudo-feature vectors for previous classes to mitigate catastrophic forgetting, and it also learns the most informative samples in each increment. As expected, FT suffers from catastrophic forgetting since it finetunes the network on new data in each increment. Regularization techniques (LWF and EWC) also suffer from catastrophic forgetting and produce the same accuracy as FT. CWR and iCaRL produce slightly better accuracy, however they also suffer from catastrophic forgetting. NCM produces much better accuracy than the other approaches with the help of centroids to remember past classes and prioritize objects to learn. However, NCM's accuracy is significantly inferior to FLB and GBCL. CBCL produces the best accuracy among the SOTA methods because it learns multiple centroids per class. However, CBCL’s accuracy is still lower than GBCL. 

Note that after 400 increments, FLB produces $\sim$89\% final accuracy and GBCL produces $\sim$88\% (1\% lower) accuracy. The reason is that after 400 increments both models learned all the training samples in the dataset, and therefore the advantage of informative sampling in GBCL fades away. Further, unlike FLB, GBCL does not have access to all the data in the previous increments, and it suffers from slight forgetting compared to batch learning (FLB).  

In terms of memory storage, GBCL stores 239 clusters/Gaussian distributions (centroids and covariance matrices) for all the classes. To avoid huge memory storage, we only use the diagonal entries of the covariance matrix for all the experiments. Therefore, GBCL stores 239$\times$2=478 vectors of size 512 in memory after learning over 70 increments on the CORe-50 dataset, which requires only 0.97 MB. In contrast, FLB (batch learning approach) stores all 21000 features vectors for all the classes, which requires 43.08 MB (44 times more than GBCL). For other approaches, FT does not store any data, while LWF, EWC store minimal information, such as the predicted labels for the previous classes and fisher matrix to capture weight importance. CWR stores an extra weight matrix for the classification layer which requires 0.02 MB of storage space. iCaRL stores 2000 raw images for the previous classes, which requires 393 MB. NCM stores only 10 centroids for the previous classes, which requires 0.02 MB. Finally, CBCL stores 315 centroids which requires 0.64 MB. Note that all of the approaches also store a ResNet-18 model trained on the previous classes which requires 83 MB of space. This analysis shows that GBCL provides the best trade-off between memory storage and the overall performance in comparison with the other approaches.

\subsubsection{Comparison with Random Sampling}
To provide more insight into our approach, we tested our GBCL approach without active learning (GBCL-random). In this case, GBCL gets a random object to be labeled in each increment. GBCL-random produced significantly lower accuracy than FLB and GBCL for all the increments, especially in earlier increments. These results depict the contribution of our active learning approach to efficiently choose which objects to learn. In comparison with other SOTA approaches, GBCL-random beats all the approaches (except CBCL) in later increments but it is below NCM and CBCL in earlier increments. These results also demonstrate the effectiveness of GBCL (Section \ref{sec:gbcl}) of our approach to mitigate catastrophic forgetting.

\begin{table}
\centering
\small
\begin{tabular}{ |P{2.0cm}|P{.55cm}|P{.55cm}|P{.55cm}|P{0.55cm}|P{.55cm}|P{.55cm}|}
     \hline
    \textbf{Increments} & 40 & 80 & 120 & 160 & 200 & 240 \\
    \hline
    \textbf{Accuracy (\%)} & 56.7 & 76.5 & 84.1 & 87.5 & 87.9 & 88.3\\
    \hline
    \textbf{No. of Classes} & 18 & 20 & 20 & 20 & 20 & 20\\
 \hline
 \end{tabular}
 \caption{Test set accuracy and number of classes learned by Pepper over 240 increments.} 
 \label{tab:pepper_exp}
 \end{table}

\subsubsection{Comparison of Learning Efficiency}
To further evaluate the efficiency of of our approach for FoCAL, we report the total number of increments taken by all the approaches to learn all 10 classes (see Table \ref{tab:classes}). GBCL requires the least number of increments (15) to learn the 10 classes. FLB takes 19 increments to learn all the 10 classes because it uses the softmax based sampling. NCM and iCaRL also take similar number of increments because they use centroids to calculate the uncertainty scores but do not have access to all the data of the previous increments. CBCL is the closest to our approach and learns all the classes in 17 increments. FT, LWF and EWC all use softmax-based sampling without using any previous data, so they take a large number of increments to learn all 10 classes. Lastly, random sampling takes the most number of increments (except for FT, LWF, and EWC) to learn all the 10 classes. These results clearly show that our approach is the most efficient at finding the unknown objects and learning them in a small number of increments.

\subsection{Experiment on the Pepper Robot}
In this experiment we used the Pepper robot to learn 240 household objects belonging to 20 classes (12 objects per class). We performed the experiment in an indoor lab environment with objects placed at four different locations in the lab with different backgrounds. At each location we placed 5 different objects on a table (Figure \ref{fig:clutter_images} (a, b)). For this setup, the robot first localized the 5 individual objects in the image. We use RetinaNet \cite{Lin_2017_ICCV} pre-trained on MS COCO dataset \cite{Lin14} for object detection and localization. The robot then captures images of the individual objects from multiple viewpoints (see Figure \ref{fig:clutter_images} (c) for an example). These images are used by the active learning module of GBCL to find the most informative object. The user (experimenter) then provides the class label of the chosen object to the robot using a keyboard. The newly labeled object is then used to update GBCL. Details about our complete system for the Pepper robot are in Section \ref{sec:robot_architecture}.

This experiment tried to reproduce some of the challenges that might arise in realistic household environments. For example, a large number of objects were learned at different times, over multiple days (8 weeks), depicting a true lifelong learning robot in real-world environments. Different objects were used in the training and test sets. Moreover, the background of the objects was non-ideal since many parts of the background could be viewed as objects by the object detector. For example, consider Figure \ref{fig:clutter_images} where the air-conditioner and the window could be viewed as objects by the object detector. Objects also varied in size and some were almost transparent. Further, the lighting conditions were not ideal. Some images were taken at night, while some during the day with different variations of sun light coming through the windows in the environment (see Figure 2 and Figure 6 for examples).

To evaluate this experiment, we created a test set of 60 objects (3 objects per class) that were different from objects in the training set. We tested Pepper's ability to classify the objects in the test set after 40, 80, 120, 160, 200, and 240 increments of learning new objects. We also report the number of object classes learned by Pepper after each of these increments (Table 2). After only 80 increments, Pepper learned all 20 object classes and achieved 76.5\% accuracy on the fixed test set. For the rest of the increments the increase in accuracy was minimal, especially in later increments (120-240), similar to the results on the CORe-50 dataset. The reason is that in the earlier increments Pepper learned objects belonging to a large number of classes quickly which increased the overall accuracy. However, in the later increments Pepper only learned about more objects of the previously learned classes, thus the increase in overall accuracy was minimal. We should also note that even after 240 increments (same as the total number of training objects) Pepper only learned 197 objects. The reason is that the object localization module was not perfect, and it failed to detect many objects (43 objects).

\section{Ablation Studies}
\label{sec:ablation}
We performed two ablation studies to examine the contribution of different components of our approach and the effect of hyperparameter values on GBCL's performance. This set of experiments were performed on the dataset collected by the Pepper robot (we call it Pepper dataset in this paper). As GBCL's accuracy saturates by 160 increments (Table \ref{tab:pepper_exp})
, we performed all of these experiments for 160 increments, with a single object learned in each increment. We report accuracy for increments 0, 40, 60, 80, 120, 140, 160, and the average incremental accuracy. The rest of the experimental setup was the same as in the experiments for the CORe-50 dataset (Section \ref{sec:core50_experiment}).  

\begin{figure}
\centering
\includegraphics[width=1.0\linewidth]{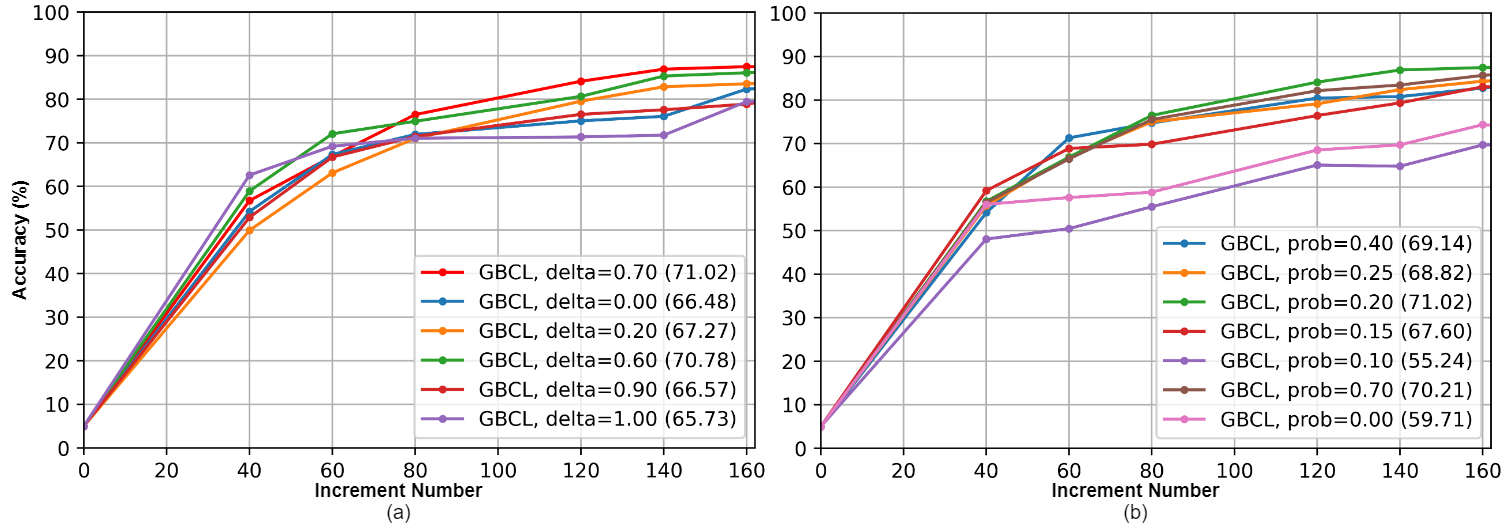}
\caption{\small The effects of varying $\delta$, and the probability threshold $P$ on the classification accuracy on the test set of Pepper dataset. While changing one of the two parameters, the other parameter's values was set to be $\delta=0.7$, $P=0.2$ for the best results}
\label{fig:ablation_figure}
\end{figure}

For the first ablation study, we investigate the effect of hyperparameter $\delta$ on choosing the two acquisition functions (entropy and viewpoint consistency) for active learning. Figure \ref{fig:ablation_figure} (a) depicts the impact of choosing different values ranging from 0 to 1, for the hyperparameter $\delta$. For the extreme values i.e. 0 and 1, our model chooses only one of the acquisition functions for active learning. Therefore, the model produces the lowest accuracy for $\delta$ values 0 and 1. However, there is no significant difference in accuracy when choosing either of the acquisition functions. For other $delta$ values close to 0 or 1, the accuracy increases slightly, but the best accuracy is achieved with $\delta=0.7$. However, for other values of $\delta$ close to 0.7 (such as 0.6), there is not a significant difference in accuracy (only 0.07\%), which shows that our approach is not highly sensitive to a large range of possible values of $\delta$. These results also confirm that choosing a combination of the two acquisition functions produces better performance than choosing either of them alone.

For the second ablation study, we investigate the effect of hyperparameter $P$ that determines the number of clusters (or Gaussian distributions) generated for each object class. Figure \ref{fig:ablation_figure} (b) shows the impact of $P$ on GBCL's performance. For $P=0$, GBCL simply stores a single cluster for each class which might not be enough to capture the complex distributions of the object classes. Therefore, for $P$ values close to 0, there is a significant drop in GBCL's accuracy ($\sim$18\%). As the $P$ value increases, the model starts to assign more clusters for each object class, and thus it is able to capture the complex distributions of the object data. The average incremental accuracy of the model increase significantly when $P=0.15$ with 41 total clusters for all 20 object classes. The best accuracy is achieved for $P=0.2$ (83 total clusters), however, for a large range of $P$ values the average incremental accuracy stays within a range of $\sim$3\%. As the $P$ value continues to increase, the model starts to assign more clusters per class, therefore for higher values of $P$ the model uses more storage space. Further, for $P$ values close to $1$, the model starts to recruit more clusters per class with only a few features (or a single feature) assigned to each cluster. Thus, the model starts to resemble more with the batch learning case when all the features of the previously learned classes can be stored. Because of this, we start to see an increase in accuracy again when $P=0.7$ (379 clusters, 4.5 times more than the number of clusters stored for $P=0.2$). Overall, these results show that our model is not highly sensitive to probability threshold values within a large range, and values close to $P=0.2$ provide a nice trade-off between memory storage and model's performance.  

Finally, note that we found both hyperparameters using cross-validation on the CORe-50 dataset and not on the Pepper dataset. However, even on the Pepper dataset the same hyperparameter values produce the best results. This shows that the chosen hyperparameter values are not dependent on a single dataset.

\section{Related Work}
\label{sec:related_work}
\noindent Most continual learning approaches are designed for class-incremental setting in which a model (often a neural network) learns from training data of different classes in different increments and then evaluated on the test data of all the classes learned so far \cite{Rebuffi_2017_CVPR}. Most existing class-incremental learning (CIL) methods avoid catastrophic forgetting by storing a portion of the training samples from previous classes and retraining the model on a mixture of the stored data and new data \cite{Rebuffi_2017_CVPR,Castro_2018_ECCV}. However, this approach does not scale as additional data exhausts memory capacity limiting performance in real-world applications. To avoid this problem, some CL approaches use regularization techniques \cite{Li18,kirkpatrick17}. Although these approaches solve the memory storage issues, their performance is significantly inferior to approaches that store old class data. A few works have been proposed for FSCIL, to continually learn from a few-examples per class using limited memory \cite{Ayub_2020_CVPR_Workshops,Tao_2020_CVPR,tao_2020_ECCV,ayub_icra_2021}. However, as mentioned in \cite{mudt2020,mundt2022clevacompass}, one significant limitation of all CL approaches is that they require the complete training data of each class to be labeled and available in a single increment, however in real-world robotic applications data is available in a streaming manner (online learning \cite{Aljundi_2019_CVPR}) and it might be mostly unlabeled. 


For object learning, many researchers have presented active learning techniques using uncertainty sampling \cite{Beluch_2018_CVPR,Gal17,Siddiqui_2020_CVPR,Yoo_2019_CVPR,Shen_2019_ICCV,singh20}. Most of these approaches train deep neural networks with special loss terms such that the networks can predict the most uncertain samples. All of these approaches, however, are trained for batch learning setting and will thus suffer from catastrophic forgetting when attempting to learn continually. Further, active learning approaches can predict unknown classes in the first increment after batch training. However, in a continual learning setting these approaches lose their ability to recognize unknown classes in subsequent increments because the model will assign the unknown classes to the newly learned classes (learned in the previous increments) that were unknown in the first increment. 

Another related field to active learning is open-set recognition (or out of the distribution detection (OOD)) \cite{boult2019learning,Bendale_2015_CVPR,Bendale_2015_CVPR,lee2018simple}, in which a model that is trained on a set of tasks/classes might face samples belonging previously unknown classes in the test set. Therefore, during the testing/deployment phase, the model must be able to detect which of the samples belong to unknown classes. Although methods for OOD \cite{scheirer2014probability,yoshihashi2019classification} have been developed to detect unknown classes, these models do not train on the newly encountered unknown samples. Mundt et al. \cite{mudt2020} present a thorough analysis of previous continual and active learning, and OOD works, and suggest a common viewpoint with open-set recognition acting as a natural interface between continual and active learning. Towards this goal, we present an approach that allows an AI agent to actively select the most informative (unknown) samples from unlabeled data, and continually learn from the actively selected object samples.     

\section{Conclusion}
\label{sec:conclusion}

This paper presents and evaluates a novel method for a challenging problem: Few-Shot Continual Active Learning (FoCAL) for object classification. Experimental results demonstrate that our approach for FoCAL is highly efficient and helps the model learn the most uncertain objects continually without forgetting earlier classes. Our approach not only outperforms the SOTA approaches on a benchmark dataset but also allows a robot to actively learn objects in a real environment over a long period of time. Finally, we 
have also released the dataset collected by the Pepper robot 
as a benchmark for future evaluations of FoCAL models. 

The work presented here has a few limitations. 1) The experimental setup is still quite simple compared to a real household environment. 2) We assume that correct object labels are provided by the human assistant to the robot. 3) We did not recruit real participants to teach the robot, but the experimenter acted as a user for the robot. 4) We used a fixed feature extractor and object detector. In the future, we hope to work on these limitations and test our system with real human participants. Particularly, we hope to develop techniques that can allow feature representations to be learned from a few samples over a large number of increments. 

There are a number of positive and negative long-term outcomes of this work. 
With respect to positive outcomes, the improved performance for continual learning by our approach may one day lead to robots that can tailor their behavior and learning to the needs of a person. We envision a home service robot \cite{ayub_icdl_2021} that uses continual learning to learn a person's food preferences before serving them breakfast, or a domestic service robot that learns how a person prefers their home to be cleaned and adjusts its behavior to their preferences.    
In terms of negative outcomes, one potential downside 
is that an active learning robot might learn unnecessary objects which are never going to be used by the robot. Learning such objects can potentially 
decrease the recognition performance on necessary objects that are required by the robot. Using FoCAL to learn unnecessary objects can also take valuable time that might be better used by the robot to perform necessary actions. For example, consider a dish washing robot which only needs to know kitchen utensils. The robot does not need to learn mechanical tools or children's toys but uncertainty-based sampling will cause the robot to learn these unnecessary objects. In such cases, some constraints might be added to a FoCAL system in order to avoid the robot from learning unnecessary objects.


\section*{Acknowledgements}
The authors acknowledge helpful comments of Alan R. Wagner for a preliminary version of this work.

{\small
\bibliographystyle{unsrt}
\bibliography{main.bib}

\begin{thebibliography}{10}

\bibitem{ayub_icra_workshop_2021}
Ali Ayub and Alan~R. Wagner.
\newblock Learning novel objects continually through curiosity.
\newblock In {\em IEEE ICRA 2021 (Workshop titled, Towards Curious Robots:
  Modern Approaches for Intrinsically-Motivated Intelligent Behavior))}, 2021.

\bibitem{ayub_roman_workshop_2021}
Ali Ayub and Alan~R. Wagner.
\newblock Online learning of objects through curiosity-driven active learning.
\newblock In {\em IEEE RoMan (Workshop on Lifelong Learning for Long-term
  Human-Robot Interaction)}, 2020.

\bibitem{Rebuffi_2017_CVPR}
Sylvestre-Alvise Rebuffi, Alexander Kolesnikov, Georg Sperl, and Christoph~H.
  Lampert.
\newblock i{C}a{RL}: Incremental classifier and representation learning.
\newblock In {\em The IEEE Conference on Computer Vision and Pattern
  Recognition (CVPR)}, July 2017.

\bibitem{kirkpatrick17}
James Kirkpatrick, Razvan Pascanu, Neil~C. Rabinowitz, Joel Veness, Guillaume
  Desjardins, Andrei~A. Rusu, Kieran Milan, John Quan, Tiago Ramalho, Agnieszka
  Grabska-Barwinska, Demis Hassabis, Claudia Clopath, Dharshan Kumaran, and
  Raia Hadsell.
\newblock Overcoming catastrophic forgetting in neural networks.
\newblock {\em Proceedings of the National Academy of Sciences of the United
  States of America}, 114(13):3521--3526, 2017.

\bibitem{Ayub_IROS_20}
Ali Ayub and Alan~R. Wagner.
\newblock Tell me what this is: Few-shot incremental object learning by a
  robot.
\newblock {\em IEEE/RSJ International Conference on Intelligent Robots and
  Systems (IROS)}, 2020.

\bibitem{ayub2021eec}
Ali Ayub and Alan Wagner.
\newblock {EEC}: Learning to encode and regenerate images for continual
  learning.
\newblock In {\em International Conference on Learning Representations}, 2021.

\bibitem{Hou_2019_CVPR}
Saihui Hou, Xinyu Pan, Chen~Change Loy, Zilei Wang, and Dahua Lin.
\newblock Learning a unified classifier incrementally via rebalancing.
\newblock In {\em The IEEE Conference on Computer Vision and Pattern
  Recognition (CVPR)}, June 2019.

\bibitem{Wu_2019_CVPR}
Yue Wu, Yinpeng Chen, Lijuan Wang, Yuancheng Ye, Zicheng Liu, Yandong Guo, and
  Yun Fu.
\newblock Large scale incremental learning.
\newblock In {\em The IEEE Conference on Computer Vision and Pattern
  Recognition (CVPR)}, June 2019.

\bibitem{Tao_2020_CVPR}
Xiaoyu Tao, Xiaopeng Hong, Xinyuan Chang, Songlin Dong, Xing Wei, and Yihong
  Gong.
\newblock Few-shot class-incremental learning.
\newblock In {\em Proceedings of the IEEE/CVF Conference on Computer Vision and
  Pattern Recognition (CVPR)}, June 2020.

\bibitem{Ayub_2020_CVPR_Workshops}
Ali Ayub and Alan~R. Wagner.
\newblock Cognitively-inspired model for incremental learning using a few
  examples.
\newblock In {\em The IEEE/CVF Conference on Computer Vision and Pattern
  Recognition (CVPR) Workshops}, June 2020.

\bibitem{Dehghan19}
M.~Dehghan, Z.~Zhang, M.~Siam, J.~Jin, L.~Petrich, and M.~Jagersand.
\newblock Online object and task learning via human robot interaction.
\newblock In {\em 2019 International Conference on Robotics and Automation
  (ICRA)}, pages 2132--2138, May 2019.

\bibitem{ayub_icsr_2020}
Ali Ayub and Alan~R. Wagner.
\newblock What am i allowed to do here?: Online learning of context-specific
  norms by pepper.
\newblock In Alan~R. Wagner, David Feil-Seifer, Kerstin~S. Haring, Silvia
  Rossi, Thomas Williams, Hongsheng He, and Shuzhi Sam~Ge, editors, {\em Social
  Robotics}, pages 220--231, Cham, 2020. Springer International Publishing.

\bibitem{french19}
Robert~M. French.
\newblock Dynamically constraining connectionist networks to produce
  distributed, orthogonal representations to reduce catastrophic interference.
\newblock {\em Proceedings of the Sixteenth Annual Conference of the Cognitive
  Science Society}, pages 335--340, 2019.

\bibitem{Siddiqui_2020_CVPR}
Yawar Siddiqui, Julien Valentin, and Matthias Niessner.
\newblock Viewal: Active learning with viewpoint entropy for semantic
  segmentation.
\newblock In {\em Proceedings of the IEEE/CVF Conference on Computer Vision and
  Pattern Recognition (CVPR)}, June 2020.

\bibitem{Yoo_2019_CVPR}
Donggeun Yoo and In~So Kweon.
\newblock Learning loss for active learning.
\newblock In {\em Proceedings of the IEEE/CVF Conference on Computer Vision and
  Pattern Recognition (CVPR)}, June 2019.

\bibitem{singh20}
Devendra~Singh Chaplot, Helen Jiang, Saurabh Gupta, and Abhinav Gupta.
\newblock Semantic curiosity for active visual learning.
\newblock In Andrea Vedaldi, Horst Bischof, Thomas Brox, and Jan-Michael Frahm,
  editors, {\em Computer Vision -- ECCV 2020}, pages 309--326, Cham, 2020.
  Springer International Publishing.

\bibitem{mudt2020}
Martin Mundt, Yong~Won Hong, Iuliia Pliushch, and Visvanathan Ramesh.
\newblock A wholistic view of continual learning with deep neural networks:
  Forgotten lessons and the bridge to active and open world learning.
\newblock {\em CoRR}, abs/2009.01797, 2020.

\bibitem{Gal17}
Yarin Gal, Riashat Islam, and Zoubin Ghahramani.
\newblock Deep bayesian active learning with image data.
\newblock In {\em Proceedings of the 34th International Conference on Machine
  Learning - Volume 70}, page 1183–1192. JMLR.org, 2017.

\bibitem{Aljundi_2019_CVPR}
Rahaf Aljundi, Klaas Kelchtermans, and Tinne Tuytelaars.
\newblock Task-free continual learning.
\newblock In {\em The IEEE Conference on Computer Vision and Pattern
  Recognition (CVPR)}, June 2019.

\bibitem{Robins95}
Anthony Robins.
\newblock Catastrophic forgetting, rehearsal and pseudorehearsal.
\newblock {\em Connection Science}, 7(2):123--146, 1995.

\bibitem{shannon_1949}
Claude~Elwood Shannon.
\newblock {\em A mathematical theory of communication}.
\newblock University of Illinois Press, 1949.

\bibitem{lomonaco17}
Vincenzo Lomonaco and Davide Maltoni.
\newblock Core50: a new dataset and benchmark for continuous object
  recognition.
\newblock In {\em Proceedings of the 1st Annual Conference on Robot Learning},
  volume~78, pages 17--26, 2017.

\bibitem{dasgupta07}
Sanjoy Dasgupta and Daniel Hsu.
\newblock On-line estimation with the multivariate gaussian distribution.
\newblock In {\em Twentieth Annual Conference on Learning Theory}, volume 4539,
  page 278–292, June 2007.

\bibitem{torch19}
Adam Paszke, Sam Gross, Francisco Massa, Adam Lerer, James Bradbury, Gregory
  Chanan, Trevor Killeen, Zeming Lin, Natalia Gimelshein, Luca Antiga, Alban
  Desmaison, Andreas Kopf, Edward Yang, Zachary DeVito, Martin Raison, Alykhan
  Tejani, Sasank Chilamkurthy, Benoit Steiner, Lu~Fang, Junjie Bai, and Soumith
  Chintala.
\newblock Pytorch: An imperative style, high-performance deep learning library.
\newblock In {\em Advances in Neural Information Processing Systems 32}, pages
  8024--8035. Curran Associates, Inc., 2019.

\bibitem{He_2016_CVPR}
Kaiming He, Xiangyu Zhang, Shaoqing Ren, and Jian Sun.
\newblock Deep residual learning for image recognition.
\newblock In {\em The IEEE Conference on Computer Vision and Pattern
  Recognition (CVPR)}, June 2016.

\bibitem{Krizhevsky12}
Alex Krizhevsky, Ilya Sutskever, and Geoffrey~E Hinton.
\newblock Imagenet classification with deep convolutional neural networks.
\newblock In {\em Advances in Neural Information Processing Systems}, pages
  1097--1105. 2012.

\bibitem{Li18}
Z.~Li and D.~Hoiem.
\newblock Learning without forgetting.
\newblock {\em IEEE Transactions on Pattern Analysis and Machine Intelligence},
  40(12):2935--2947, Dec 2018.

\bibitem{Chen19}
Wei-Yu Chen, Yen-Cheng Liu, Zsolt Kira, Yu-Chiang~Frank Wang, and Jia-Bin
  Huang.
\newblock A closer look at few-shot classification.
\newblock In {\em International Conference on Learning Representations}, 2019.

\bibitem{Lin_2017_ICCV}
Tsung-Yi Lin, Priya Goyal, Ross Girshick, Kaiming He, and Piotr Dollar.
\newblock Focal loss for dense object detection.
\newblock In {\em The IEEE International Conference on Computer Vision (ICCV)},
  Oct 2017.

\bibitem{Lin14}
Tsung-Yi Lin, Michael Maire, Serge~J. Belongie, James Hays, Pietro Perona, Deva
  Ramanan, Piotr Doll{\'a}r, and C.~Lawrence Zitnick.
\newblock Microsoft coco: Common objects in context.
\newblock In {\em ECCV}, 2014.

\bibitem{Castro_2018_ECCV}
Francisco~M. Castro, Manuel~J. Marin-Jimenez, Nicolas Guil, Cordelia Schmid,
  and Karteek Alahari.
\newblock End-to-end incremental learning.
\newblock In {\em The European Conference on Computer Vision (ECCV)}, September
  2018.

\bibitem{tao_2020_ECCV}
Xiaoyu Tao, Xinyuan Chang, Xiaopeng Hong, Xing Wei, and Yihong Gong.
\newblock Topology-preserving class-incremental learning.
\newblock In {\em Computer Vision – ECCV 2020: 16th European Conference,
  Glasgow, UK, August 23–28, 2020, Proceedings, Part XIX}, page 254–270.
  Springer-Verlag, 2020.

\bibitem{ayub_icra_2021}
Ali Ayub and Alan~R. Wagner.
\newblock F-siol-310: A robotic dataset and benchmark for few-shot incremental
  object learning.
\newblock In {\em 2021 IEEE International Conference on Robotics and Automation
  (ICRA)}, pages 13496--13502, 2021.

\bibitem{mundt2022clevacompass}
Martin Mundt, Steven Lang, Quentin Delfosse, and Kristian Kersting.
\newblock {CLEVA}-compass: A continual learning evaluation assessment compass
  to promote research transparency and comparability.
\newblock In {\em International Conference on Learning Representations}, 2022.

\bibitem{Beluch_2018_CVPR}
William~H. Beluch, Tim Genewein, Andreas Nürnberger, and Jan~M. Köhler.
\newblock The power of ensembles for active learning in image classification.
\newblock In {\em Proceedings of the IEEE Conference on Computer Vision and
  Pattern Recognition (CVPR)}, June 2018.

\bibitem{Shen_2019_ICCV}
Tingke Shen, Amlan Kar, and Sanja Fidler.
\newblock Learning to caption images through a lifetime by asking questions.
\newblock In {\em Proceedings of the IEEE/CVF International Conference on
  Computer Vision (ICCV)}, October 2019.

\bibitem{boult2019learning}
Terrance~E Boult, Steve Cruz, Akshay~Raj Dhamija, Manuel Gunther, James
  Henrydoss, and Walter~J Scheirer.
\newblock Learning and the unknown: Surveying steps toward open world
  recognition.
\newblock In {\em Proceedings of the AAAI conference on artificial
  intelligence}, volume~33, pages 9801--9807, 2019.

\bibitem{Bendale_2015_CVPR}
Abhijit Bendale and Terrance Boult.
\newblock Towards open world recognition.
\newblock In {\em The IEEE Conference on Computer Vision and Pattern
  Recognition (CVPR)}, June 2015.

\bibitem{lee2018simple}
Kimin Lee, Kibok Lee, Honglak Lee, and Jinwoo Shin.
\newblock A simple unified framework for detecting out-of-distribution samples
  and adversarial attacks.
\newblock {\em Advances in neural information processing systems}, 31, 2018.

\bibitem{scheirer2014probability}
Walter~J Scheirer, Lalit~P Jain, and Terrance~E Boult.
\newblock Probability models for open set recognition.
\newblock {\em IEEE transactions on pattern analysis and machine intelligence},
  36(11):2317--2324, 2014.

\bibitem{yoshihashi2019classification}
Ryota Yoshihashi, Wen Shao, Rei Kawakami, Shaodi You, Makoto Iida, and Takeshi
  Naemura.
\newblock Classification-reconstruction learning for open-set recognition.
\newblock In {\em Proceedings of the IEEE/CVF Conference on Computer Vision and
  Pattern Recognition}, pages 4016--4025, 2019.

\bibitem{ayub_icdl_2021}
Ali Ayub, Chrystopher~L. Nehaniv, and Kerstin Dautenhahn.
\newblock Don't forget to buy milk: Contextually aware grocery reminder
  household robot.
\newblock {\em arXiv:2207.09050}, 2022.

\bibitem{Krizhevsky09}
Alex Krizhevsky.
\newblock Learning multiple layers of features from tiny images, 2009.
\newblock Technical report, University of Toronto.

\bibitem{Hinton15}
Geoffrey Hinton, Oriol Vinyals, and Jeffrey Dean.
\newblock Distilling the knowledge in a neural network.
\newblock In {\em NIPS Deep Learning and Representation Learning Workshop},
  2015.

\end{thebibliography}
}

\appendix
\section{CORe-50 Dataset}
\label{sec:core50}
CORe-50 is an object recognition dataset that was developed by Lomonaco et al. \cite{lomonaco17} for continual learning that contains 50 household objects belonging to 10 classes (5 objects per class). Classification can be performed at the object level (50 classes) or category level (10 classes). Each object class is captured with 11 different background environments from 11 different recording sessions. Each session (video recording) is composed of an approximately 15 second video clip recorded at 20 fps. We used the cropped 128×128 images with sessions 3, 7, and 10 as the test set with the rest being used for training. We performed experiments at the category level with 10 classes.

\section{Comparison with AL and CL Setups}
\label{sec:comparison_with_setups}
For FoCAL experiments on the CORe-50 dataset, we develop a new experimental setup, as the standard active learning and continual learning experimental setups are not sufficient to test FoCAL. In a standard active learning (AL) experiment \cite{Beluch_2018_CVPR}, the training set of a large dataset (such as CIFAR-10 \cite{Krizhevsky09}) is divided into smaller subsets ($\sim$10000 unlabeled images), where each subset can contain images belonging to all the classes in the dataset. In a single increment, an active learning model is provided with a subset and the model chooses $k$($\sim$1000) instances to be labeled. After each increment, the model is tested on the complete test set of the dataset. Note that in subsequent increments, the model still has access to the previous data (batch learning). On the other hand, in a standard continual learning (CL) experiment \cite{Ayub_2020_CVPR_Workshops,Rebuffi_2017_CVPR}, a large dataset is divided into smaller subsets where each subset contains complete training set of a subset of the classes. In a single increment, a continual learning model is provided with a subset with the ground truth labels. After training in each increment, the model is tested only on the test set of the classes learned so far. Also, in the subsequent increments, the model does not have access to the previous data subsets.  

In contrast, for FoCAL experiments on CORe-50 we combined all 8 training sessions in CORe-50 to generate 400 training object instances. In each increment, we randomly sampled $m=5$ object instances (from 400 instances) and allowed the model to learn the label of $k=1$ out of 5 object instances, making it a challenging problem to learn from a single object instance in each increment. Once an object was learned by the model, it was removed from the complete set of objects and, thus, was not available to the model in later increments. Hence, we allowed the model to learn all objects in 400 increments with one object learned in each increment. 

\section{SOTA CL Approaches}
\label{sec:sota_cl_approaches}
We compare our approach against 6 continual learning approaches (LWF \cite{Li18}, EWC \cite{kirkpatrick17}, CWR \cite{lomonaco17}, iCaRL \cite{Rebuffi_2017_CVPR}, NCM \cite{Dehghan19}, CBCL \cite{Ayub_2020_CVPR_Workshops}), finetuning (FT) and a few-shot batch learning baseline (FLB) \cite{Chen19}. FLB uses a pre-trained CNN to extract feature vectors for images and trains a linear classifier using cross entropy loss. FLB is trained on the complete training data of the previous increments and the current increment. In other words, FLB does not learn continually and therefore should have an advantage over other continual learning approaches (including ours). FT uses the same architecture as FLB but FT is trained only on the data of the current increment. FT suffers from catastrophic forgetting and therefore should produce lower accuracy than other continual learning approaches. Nearest Class Mean (NCM) classifier computes a single centroid for each class as the mean of all the feature vectors of the images in the training set for each class. To predict the label for a test image, NCM assigns it the class label of the closest centroid. NCM avoids catastrophic forgetting by using centroids. Each class centroid is computed using only the training data of that class, hence even if the classes are learned continually, the centroids for previous classes are not affected when new classes are learned. Centroid-Based Concept Learning (CBCL) generalizes NCM, and uses a cognitively-inspired clustering approach to learn multiple centroids per class, instead of a single centroid. CBCL uses a k-nearest centroids approach for classification of test images. Incremental classifier and representation learning (iCaRL)~\cite{Rebuffi_2017_CVPR} combines knowledge distillation~\cite{Hinton15} and NCM for class-incremental learning. Knowledge distillation uses a distillation loss term that forces the labels of the training data of previously learned classes to remain the same when learning new classes. iCaRL uses the old class data while learning a representation for new classes and uses the NCM classifier for classification of the old and new classes. Elastic Weight Consolidation (EWC) searchs for importance weights for each parameter in the neural network for the old classes. During learning of new classes, the gradients for the importance weights are penalized to prevent a large change in importance parameters. This way the model can mitigate catastrophic forgetting. Finally, CWR uses a similar architecture as in FLB, but it does not use the data of the old classes when leanring new classes. To mitigate catastrophic forgetting, CWR keeps the weights of the linear layer in the classifier from the previous increment, and imprints them in the updated linear layer for learning new classes.

\begin{figure*}
\centering
\includegraphics[width=1.0\linewidth]{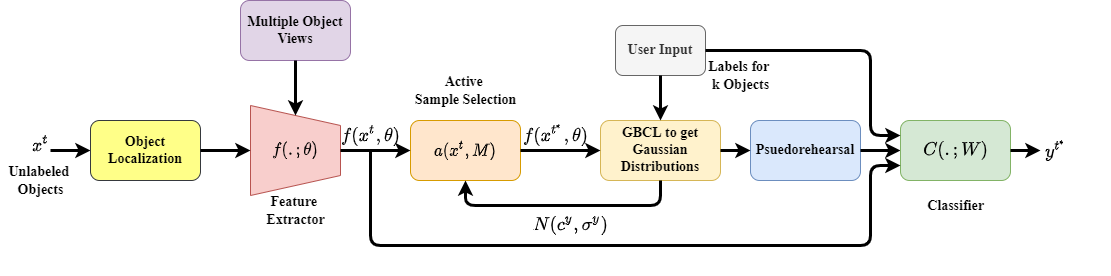}
\caption{\small Our overall framework for FoCAL. In each increment $t$, the features extracted for unlabeled objects $f(x^t;\theta)$ are passed through the acquisition function $a(x^t,\mathcal{M})$ to get $k$ most informative samples $x^{t^*}$, which are labeled by the oracle. The labeled feature vectors are used to update the GMM representation of the learned classes $Y^t$. Pseudo-rehearsal is used to replay old class data, and the classifier model $C(.;W)$ is trained on the pseudo-samples of the old classes and the labeled feature vectors in the $t$th increment.
}
\label{fig:framework_pepper}
\end{figure*}

\section{FoCAL System on a Humnoid Robot}
\label{sec:robot_architecture}
For real world robotics applications, an autonomous robot might not have perfect images of individual objects available. In such cases, the robot must detect and localize individual objects, capture multiple views of the objects itself, and then ask its human user to provide the label for the most informative objects. Therefore, we develop a complete system for our FoCAL framework (see Figure \ref{fig:framework_pepper}) that can be integrated on a real robot. The subsections below describe the different components of our framework:

\begin{figure}
\centering
\includegraphics[width=1.0\linewidth]{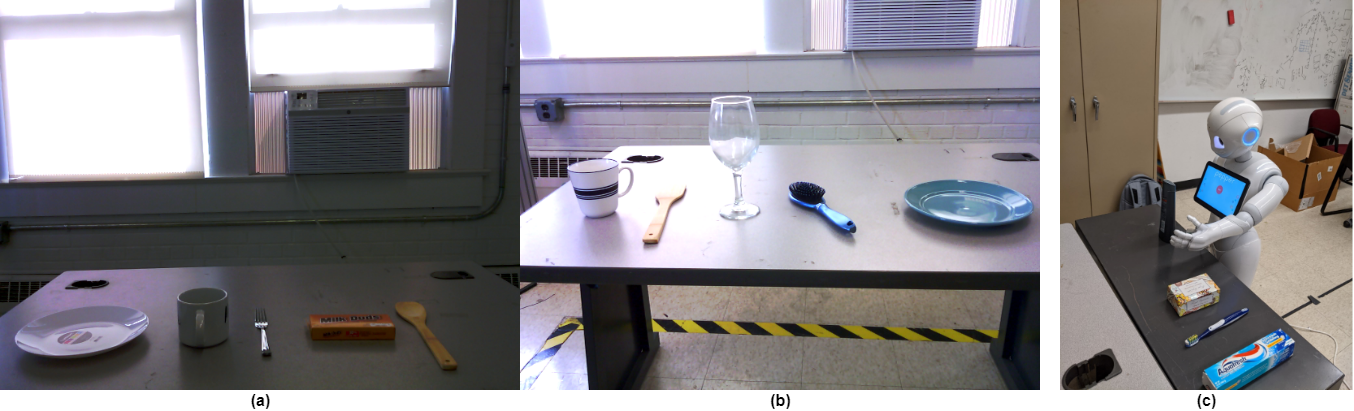}
\caption{\small (a, b) Examples of different sets of objects present on a table in different lighting conditions. Note the variations in camera angle, background and arrangements of objects in the two images. (c) Pepper robot capturing different views of an object of class shampoo using its hand camera.}
\label{fig:clutter_images}
\end{figure}

\begin{figure}
\centering
\includegraphics[width=0.65\linewidth]{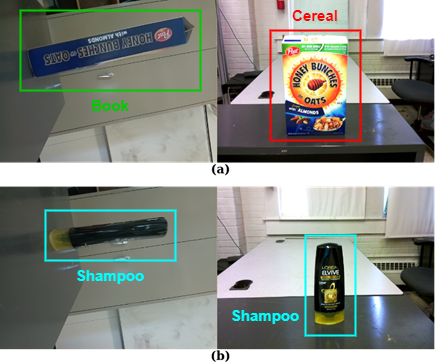}
\caption{\small \textbf{Viewpoint Consistency:} Examples of multiple views of object classes (a) cereal and (b) shampoo. (a) Inconsistent predictions for different views of the same object generates a higher reward for acquiring the label of the object, while (b) consistent predictions generate a lower reward.}
\label{fig:multiple_views}
\end{figure}

\subsection{Object Localization}
\label{sec:localization}
In a real-world environment, the robot might encounter multiple objects at a single location. Before finding the uncertainty scores for unlabeled objects, the robot must first detect, localize, and get images of individual objects from a set of objects (see Figure \ref{fig:clutter_images} (a, b) for examples). In our framework, 
the robot uses RetinaNet \cite{Lin_2017_ICCV} for object detection and localization. This network proposes image regions likely to contain objects. After passing the image through the RetinaNet, the locations of the objects are passed on to the manipulation and image capturing module to get images of individual objects.

\subsection{Manipulation and Image Capturing Module}
\label{sec:manipulation_module}
As shown by Lomonaco et al. \cite{lomonaco17}, multiple continuous views of different objects help a model develop a better representation which improves the recognition performance of the model. In the work proposed by Lomonaco et al. \cite{lomonaco17}, a human holds an object in front of a camera and moves the object in his/her hand to get multiple views of the object. However, in real-world scenarios, an autonomous robot does not have access to unlimited human assistance. Thus, a robot must capture different views of an object autonomously. In this paper, we use the humanoid robot Pepper for the experiments. The head cameras of the Pepper robot, however, are not sufficient to capture the side views of the objects on a table. Hence, we attached a Raspberry Pi V2 camera to one of the robot's hands, so that the robot could move its hand around the objects to capture different views.

To allow the robot to capture different views of each object autonomously, we use the locations of the objects (from the object localization module) to allow the robot to move itself in front of the objects. We then allow the robot to move its arm around each object and capture the side views of the object using the camera on its hand. The objects were placed at the same height as Pepper's arms, so that same arm motion was used to capture images for all objects. We also use the head camera to capture the front/top views of the object. Figure 2 in the main paper shows examples of the images captured by Pepper's head camera and the hand camera. The arms were controlled using inverse kinematics module on the robot, where the robot moved the arms in a pre-defined motion in front of each object. 
Note that the robot could use its hand to grasp and move the object and capture images using its head camera. However, grasping and manipulating the object is a challenging problem and was not pursued as a part of this work.

After capturing multiple views of individual objects, the robot uses the AL module of our approach to find the most informative objects. The robot then asks the human user (experimenter) to provide the label of the object using the text-to-speech API. The user then provides the labels of the most informative objects as text input using a keyboard. Our system then updates the GMM representations and trains the classifier model on the newly labeled data. Details about our framework are in the main paper.

\subsection{FoCAL Dataset from the Pepper Robot}
As a part of the paper, we provide the complete dataset of object images captured by the Pepper robot (available here: \url{https://tinyurl.com/2vuwv8ye}). The complete dataset consists of 16988 images for 300 different objects belonging to 20 classes. We use 240 objects (13827 images) in the training set and 60 objects (3161 images) in the test set. 

\begin{figure}
\centering
\includegraphics[width=1.0\linewidth]{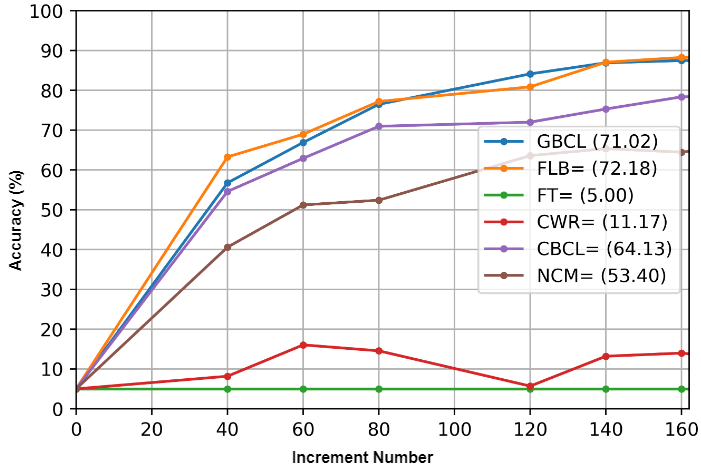}
\caption{\small Comparison of our method (blue curve) to SOTA approaches in terms of classification accuracy on the Pepper dataset. Average incremental accuracy is reported in parenthesis.}
\label{fig:pepper_dataset_exp}
\end{figure}

\section{Experiment on the Pepper Dataset}
\label{sec:pepper_dataset_experiment}
We performed further experiments on the dataset collected by the Pepper robot, and compared GBCL with other approaches. For this experiment, we only report the accuracy for 160 increments, as the final accuracy for all the models starts to saturate. Figure \ref{fig:pepper_dataset_exp} shows the comparison of GBCL with other approaches on the Pepper dataset. Similar to the results on CORe-50 dataset, GBCL is the closest to the batch learning approach (FLB) in terms of the final accuracy (less than 1\% lower after 160 increments). CBCL is the next best approach, followed by NCM. As expected, FT and CWR suffer from catastrophic forgetting. These results again confirm the effectiveness of our approach for FoCAL.

\section{Few-Shot Class Incremental Learning (FSCIL) Experiment}
\label{sec:fscil_exp}
One of the limitations of our GBCL approach is that it uses a pre-trained feature extractor. However, there are some approaches for few-shot class incremental learning (FSCIL) that also continue to update the CNN representation from only a few examples per class. An argument can be made that such approaches might be better suited for FoCAL than GBCL as they also continue to adapt the feature representation. To further explore this idea, we performed a separate experiment with GBCL to compare against the FSCIL approaches. As the FSCIL approaches are not designed for FoCAL, for a fair compairson we tested GBCL for the standard FSCIL setup \cite{Tao_2020_CVPR} on the CIFAR-100 dataset \cite{Krizhevsky09}. 

\begin{figure}
\centering
\includegraphics[width=1.0\linewidth]{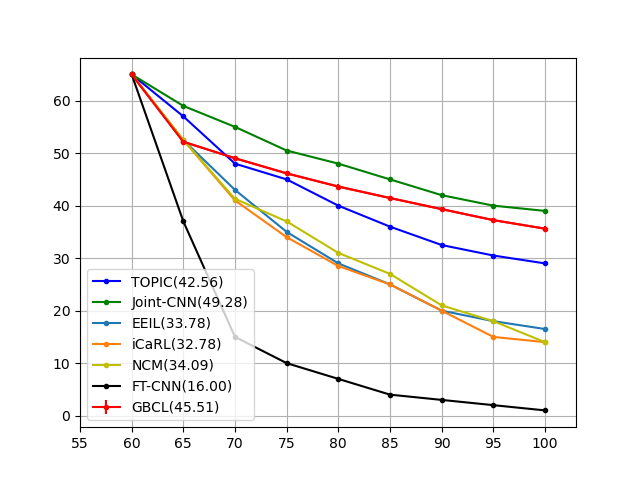}
\caption{\small Comparison of GBCL (red curve) to other approaches in terms of classification accuracy on the CIFAR-100 dataset. Average incremental accuracy is reported in parenthesis. GBCL's curve shows average and standard deviation of 3 runs with random seeds.}
\label{fig:fscil_cmparisons}
\end{figure}

In the FSCIL setup \cite{Tao_2020_CVPR}, we first train a ResNet-18 from scratch on 60 out of 100 classes of the CIFAR-100 dataset. After training on the base classes, we freeze this network and use it as a frozen feture extractor in the next increments. After the base class training, the rest of the 40 classes are learned in 8 increments with 5 classes per increment. In the spirit of FSCIL, the 60 base classes are trained with 500 images per class, while the rest of the 40 classes are trained only with 5 images per class. Further, as all the images are labeled in each increment, we remove the active learning phase from GBCL and only use the GMM clustering and pseudo-rehearsal phases for this experiment. The training settings and the probability threshold hyperparameter were kept the same as in FoCAL experiments.

Figure \ref{fig:fscil_cmparisons} shows a comparison of GBCL (red curve) with batch learning upperbound (green curve), TOPIC \cite{Tao_2020_CVPR} (blue curve) and other approaches on the CIFAR-100 dataset. Results for all the other approaches are reported from \cite{Tao_2020_CVPR}. In the first increment, all the approaches achieve similar accuracy as they all train a ResNet-18 for 60 classes in the CIFAR-100 dataset. In the second increment, GBCL's accuracy drops significantly in comparison with TOPIC, because TOPIC updates the feature representation and adapts it to the classes learned in the second increment. In contrast, GBCL uses the CNN as a fixed feature extractor and therefore its representation is not fully adapted to the classes in the second increment. In the next increment, however, GBCL performs slightly better than TOPIC because the representation learned by TOPIC is not general anymore and it became too specific to the small number of classes learned in the previous increment. In contrast, GBCL can still use the more general features learned in the first increment from a large number of classes. Further, the GMM clustering and the pseudo-rehearsal phase allow GBCL to continue to learn the complex distribution of the new classes without significant forgetting. In the subsequent increments, the accuracy gap between GBCL and TOPIC continues to increase, and after learning 100 classes GBCL achieves $\sim$8\% higher accuracy than TOPIC. Another interesting result is that GBCL also closes the accuracy gap from the Joint-CNN (batch learning) approach over 8 increments. This shows that over a large number of increments GBCL can achieve similar performance to batch learning by continuing to learn the complex distributions of the object classes. These results show that adapting the feature representation using a few examples per class can be useful when learning for a single increment only. For a large number of increments, GBCL can achieve significantly better performance by using a fixed feature extractor with GMM representations and pseudo-rehearsal.

\end{document}